%% file: glia-arxiv-v1.tex
\documentclass{article}
\usepackage{modified_style,times}

\input{math_commands.tex}

\usepackage{hyperref}
\usepackage{url}
\usepackage[pdftex]{graphicx}

\title{A Little Confidence Goes a Long Way}

\author{John Scoville, Shang Gao, Devanshu Agrawal, \& Javed Qadrud-Din \\
Casetext, a Thomson Reuters company\\
\texttt{john.scoville@thomsonreuters.com} \\
}

\thefinalcopy 
\begin{document}

\maketitle

\begin{abstract}
   We introduce a group of related methods for binary classification tasks using probes of the hidden state activations in large language models (LLMs). Performance is on par with the largest and most advanced LLMs currently available, but requiring orders of magnitude fewer computational resources and not requiring labeled data. This approach involves translating class labels into a semantically rich description, spontaneous symmetry breaking of multilayer perceptron probes for unsupervised learning and inference, training probes to generate confidence scores (prior probabilities) from hidden state activations subject to known constraints via entropy maximization, and selecting the most confident probe model from an ensemble for prediction. These techniques are evaluated on four datasets using five base LLMs.
\end{abstract}

\section{Introduction}

Incorporating constraints based on prior knowledge remains an open problem in language modeling. We offer a mechanism to incorporate certain types of constraints, based on probes of the hidden state activations of LLMs \citep{alain2018}. We take as a starting point the Contrast-Consistent Search (CCS) method of \citet{burns2023} for unsupervised training of probe models. We treat the mutually exclusive nature of binary classes as a constraint on prior knowledge, and apply the Maximum Entropy Principle (MEP) \citep{jaynes1957pt1,jaynes1957pt2,Jaynes2003Probability} to generate prior probabilities. The end result is vastly improved LLM inference on batched binary classification problems, comparable to the best general-purpose LLMs currently available but achieved using orders of magnitude fewer computational resources. These methods can be implemented efficiently on a single consumer-grade GPU. Moreover, these methods do not require supervised fine-tuning, or even access to data labels. 

Novel contributions of this study include confidence-based unsupervised ensemble learning of neural network hidden layer probes, information-theoretic loss functions for the training of these probes, a pretraining procedure to break the permutation symmetry inherent to the probe models, and a translation step to cast arbitrary class labels into a semantically meaningful form suitable for consumption by probe models. Throughout the paper we refer to these techniques collectively as Glia, after the glial cells (glia) that provide support and constraints for neurons. The methods are evaluated on four datasets: Amazon polarity, IMDB, CUAD (Contract Understanding Atticus Dataset \citep{hendrycks2021cuad}, and Learned Hands (a legal issue identification dataset).

\section{Related Work} 

\subsection{Probe models}

Probe models were used by \citet{alain2018} to evaluate the separability of features in the hidden layer activations of deep neural networks. Single-layer neural networks were trained to learn linear projections from hidden states onto features. Hidden states were analyzed at different depths in a deep neural network, and it was found that deeper layers become progressively more linearly separable. In this study, we consider deep neural network hidden state "probes" that have multiple hidden layers and may be trained in an unsupervised manner.

Various approaches have been proposed to gauge neural networks’ confidence in their predictions \citep{Goan2020, kull2019, guo2017, ding2021, zhao2021}. In \citet{burns2023} a method called Contrast Consistent Search (CCS) is used to train probes on the hidden state activations of language models to produce confidence scores. CCS deals with batches of yes/no questions and processes pairs of contradictory statements $x_+$ and $x_-$ and their final hidden layer activations $\phi(x_+)$ and $\phi(x_-)$. For example, $x_+$ could be ''Is water wet? Yes." and $x_-$ the ''Is water wet? No." The loss function for this probe attempts to learn probabilities of $x_+$ and $x_-$ by leveraging the fact that the complementary probabilities $P(\phi(x_+))$ and $1-P(\phi(x_-))$ should sum to 1. This is accomplished by a loss function incorporating a soft constraint minimizing the distance between $P(\phi(x_+))$ and $1-P(\phi(x_-))$, and also a constraint minimizing $min(P(\phi(x_+)),P(\phi(x_-)))^2$ to encourage more confident solutions. The CCS method trains a linear layer with sigmoidal activation to generate confidence scores. CCS averages $P(\phi(x_+))$ and $1-P(\phi(x_-))$ from the probe model to obtain a probability suitable for inference. In \citet{burns2023} the output of this probe for inference exceeds the accuracy of zero-shot inference by several percentage points on average.

CCS is the starting point for the Glia methods, but are several key differences between the probes used in CCS and the probes we consider in this study. The inclusion of a final softmax layer in Glia eliminates the need for the `soft' consistency loss used in CCS. Glia replaces the CCS consistency loss with entropy maximization and modifies the confidence loss from CCS with a tunable exponent hyperparameter. Another difference is that Glia uses multilayer perceptrons to learn nonlinear projections, whereas CCS is described using single-layer perceptrons that learn a linear projection\footnote{The CCS reference implementation does include multilayer perceptrons, but these were not evaluated in \citet{burns2023}.}. In the CCS method, positive and negative examples are randomly assigned to classes. This occurs because the loss function used to train probes is symmetric with respect to permutation of $x_+$ and $x_-$, therefore labels are randomly based on the initialization of the probe model parameters. \citet{burns2023} addresses this by comparing labels post-hoc, after model evaluation, flipping labels if accuracy is below 50 percent\footnote{this should be unlikely if the sample size is large enough and model/task performance is significantly better than random chance}. Since labels are ultimately needed to make predictions, we consider this a semi-supervised approach. In contrast, we propose a symmetry-breaking pretraining procedure as a means of reducing or eliminating the need for labels. This enabling fully unsupervised zero-shot inference and nearly unsupervised few-shot inference using only few-shot example labels.

In CCS, ensembling is handled by restarting training several times and choosing the model with the minimum loss. In contrast, Glia and chooses the best model from an ensemble using a confidence criterion. Another important difference is that the CCS method performs no label translation, whereas Glia does, making it perform well on true binary classification tasks, but, like CCS, worse on more general multiple-choice tasks where labels (e.g. ``A", ``B") don't share the same meaning across all examples.

\subsection{Entropy Maximization}

The Maximum Entropy Principle (MEP) is a key concept in statistical physics and Bayesian statistics that formalizes Occam's razor\footnote{Entia non sunt multiplicanda praeter necessitatem - "Entities should not be multiplied beyond necessity"}. \citet{jaynes1957pt1, jaynes1957pt2} proposed that entropy maximization should be elevated to the level of a general statistical principle. It provides a means of ''translating prior information uniquely into a prior probability assignment" \citep{Jaynes2003Probability}. The MEP dictates that one should choose the prior probability distribution having maximum entropy subject to known constraints on prior information to arrive at prior probabilities.

The R{\'e}nyi entropy $H_\alpha(\rx)=\frac{1}{1-\alpha} \log{\sum_i P(\rx_i)^\alpha}$ \citep{Renyi1961} parameterizes a family of information measures via a parameter $\alpha$. It generalizes several information measures, including Shannon information ($ \lim{\alpha \to 1}$), Hartley information, ($\alpha=0$) and collision entropy ($\alpha=2$). The MEP is traditionally expressed using the Shannon entropy, and this arises naturally from the geometry of flat statistical manifolds \citep{Amari2000Methods}. The geometry of curved statistical manifolds similarly leads to a generalized MEP utilizing R{\'e}nyi entropy \citep{morales2021generalization}. Probe models in Glia make use of this generalized MEP, maximizing R{\'e}nyi entropy to convert hidden layer activations into prior probabilities, given constraints on prior knowledge.

\section{Method description}

\subsection{Model Overview}

\begin{figure}[t]
   \centering
   \vspace{-35pt}
   \includegraphics[width=\textwidth]{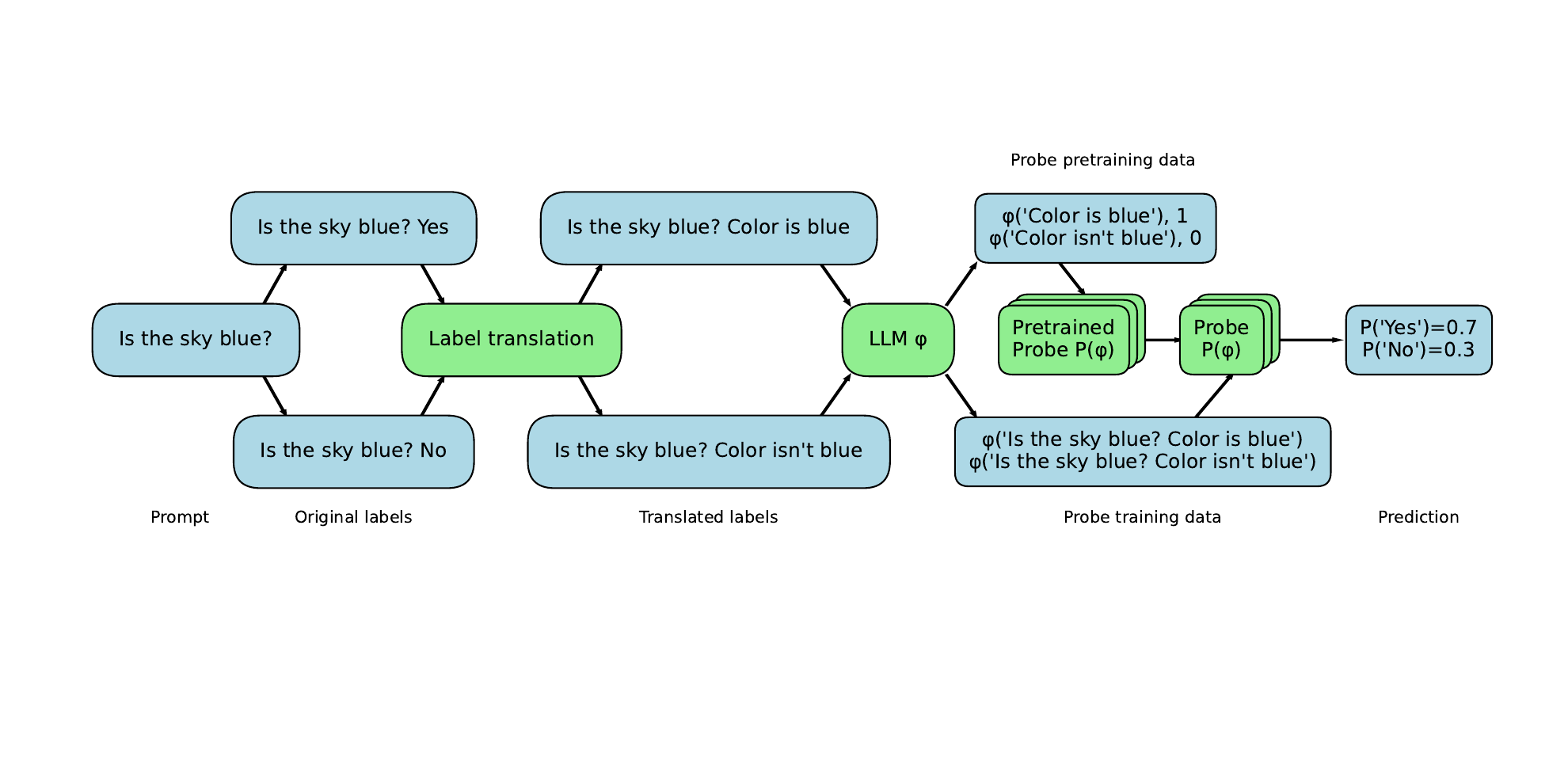}
   \vspace{-60pt}
   \caption{An illustration of the Glia framework using a single example query, ``Is the sky blue?" This represents one example from a batch that we wish to classify into one of two binary classes. First, given a task prompt (e.g. 'Is (example) blue?'), the dataset labels for the binary equivalence classes ('Yes', 'No') are translated into descriptive representations, either by a human or an instruction-tuned LLM, answering the question 'What is the meaning of a (label) response to this query?' Once descriptive labels are produced, they undergo a forward pass through an LLM. The activations of the LLMs final hidden state layer are extracted. Each of the two descriptive labels are appended to the prompt and passed through the LLM to obtain examples of each answer, again extracting the activations of the final hidden layer. Numerical labels are chosen and paired with translated label activations to create a synthetic dataset for symmetry-breaking cross-entropy pretraining. The two answers of each example become inputs to the symmetry-broken pretrained model, which is trained using the maximum entropy principle (MEP) to create a probe model. Probe pretraining and training are iterated several times to produce an ensemble of probe models. The most confident probe model from the ensemble is selected to make predictions.
   }
   \label{fig:main}
\end{figure}

The method is illustrated in Figure \ref{fig:main}. Prompts $x_+$ and $x_-$ are constructed by appending positive and negative responses to a batch of base prompts, $x$. Hidden layer activations $\phi(x_+)$ and $\phi(x_-)$ are the final hidden layer activations\footnote{Throughout this study, we refer to the activations of the final hidden state layer, unless otherwise specified, as the final layer tends to be the most separable} produced by a language model $\phi$ during a forward pass.

Hidden layer activations $\phi(x_+)$ and $\phi(x_-)$ are normalized to have zero mean and unit variance across the batch\footnote{This batch normalization differs from the layer normalization described in \citet{burns2023} but conforms with the accompanying source code}. $\phi(x_+)$ and $\phi(x_-)$ become input data for unsupervised training of an ensemble of secondary neural networks, $N$. Each of these neural networks learn an entropy-maximizing nonlinear projection of the hidden layers onto a single logit. Projections learned by maximizing entropy subject to the mutual exclusion constraint generate prior probability distributions over labels via the MEP.

Finally, for each example, $N(\phi(x_+))$ and $N(\phi(x_-))$ become inputs to binary softmax functions that normalize the probe outputs, producing the final prior probabilities that are used as confidence scores, $P(x_+) = Ce^{N(\phi(x_+))}$ and $P(x_-) = Ce^{N(\phi(x_-))}$, where $C^{-1}=e^{N(\phi(x_+))}+e^{N(\phi(x_-))}$. The softmax layer ensures that estimated prior probabilities $P(x_+)$ and $P(x_-)$ sum to unity.\footnote{Note that here, activations become inputs to the softmax layer. Linear layers before the softmax did not produce good predictions. We arbitrarily rescale the range $(Ce^{-1},Ce)$ to $(0,1)$, this does not affect predictions.}

Neural networks are trained via the Adam optimizer using hidden layers as input, without access to labels, except for examples incorporated into few-shot prompts. This procedure is repeated to train an ensemble of models, and the most confident model is selected. Finally, to perform inference, we simply check whether the prior probability scores of the most confident model are greater or less than 0.5.

\subsection{Probe Model Loss Function}

We apply the generalized MEP to the probe model loss function to produce prior probability estimates. This is accomplished via a two-part loss function, one term maximizes R{\'e}nyi entropy and the other term enforces prior knowledge about binary classification, specifically, mutual exclusion. The resulting loss function allows us to produce prior probabilities - confidence scores - using the generalized maximum entropy principle.

Let $P(x_+)$ and $P(x_-)$ be the outputs of the final softmax layer of the probe model. We introduce a constraint loss term derived from the confidence loss term used by CCS \citep{burns2023}. The CCS confidence loss is $\min{( P(x_+),P(x_-))}^2$. Since $P(x_-)=1-P(x_+)$ due to the final softmax layer, the confidence loss reduces to a function of a single variable. Additionally, we make the exponent (quadratic in CCS) a model hyperparameter, $\beta$. Our constraint loss term is:
\begin{equation}
L_{constraint}=\min{ (P(x_+),1-P(x_+))}^\beta
\end{equation}
This term of loss function encodes a known prior constraint, specifically, that the two possible labels are mutually exclusive. A peak at $p=0.5$ and minima at $p=0$ and $p=1$ make this loss function promote mutual exclusion ($P(x_+) \neq P(x_-)$) by penalizing solutions in which the outcomes have equal prior probability ($P(x_+)=P(x_-)$) and promoting solutions in which p takes on extreme values $0$ and $1$.  Other types of constraint loss functions expressing different prior knowledge or logic could potentially be used, but in this study we focus on the exclusive-or relationship of mutual exclusion in the context of binary classification.

The R{\'e}nyi entropy parameter, $\alpha$, also becomes a tunable hyperparameter of the model. Both loss terms are necessary to avoid degenerate solutions in which probabilities collapse to 0.5 and 0.5 (without the constraint loss) or to 0 and 1 (without the entropic loss), as described by \citet{burns2023}. Loss hyperparameters $\alpha=0.7$ and $\beta=0.3$ are used throughout this study\footnote{Loss hyperparameters $\alpha$ and $\beta$ are tuned with the other model hyperparameters via coordinate descent}. The full loss function used to train neural network probe models is the sum of the entropic and constraint terms. These terms can be weighted, but the model is relatively insensitive to the coefficients. Our evaluations use $c_{1}=1$ and $c_{2}=10$.
\begin{equation}
L=c_{1} L_{entropic}+ c_{2} L_{constraint}
\end{equation}

This loss function produces prior probabilities subject to known constraints on prior information, via the generalized MEP.

\subsection{Probe architecture}

The architecture of the neural network probe is a multilayer perceptron (MLP). In the evaluated implementation, this MLP has two hidden layers, the second having half the rank of the first to encourage compression while reducing parameter count. The outputs of two MLP instances (for positive and negative examples) are bridged by a softmax layer to output binary class probablities. The loss function is evaluated using these probablities. The MLP consists of an input layer of width $n$, and output layer of rank 1, and some number of hidden layers (2 hidden layers are used in our evaluation). Intermediate layer ranks vary linearly between n and 1. 

While the probe model is relatively insensitive to intermediate layer activation functions, the activation function of the final layer is significant, as it determines the range of probabilities that can be produced by the softmax layer\footnote{Sigmoidal, tanh, GeLU, SiLU, and ReLU activations were tested. ReLU activations occasionally led to model collapse (all probabilities 0.5) that was not observed with smooth activation funtions. Activation via tanh produced the best accuracy for output layers}. We use tanh activations for the final hidden layer and GeLU activations for intermediate layers.

\subsection{Ensembles of Probe Models}

Probe models are computationally inexpensive to train relative to the compute typically needed to perform inference and generate the final hidden states of a large language model. Since the overall scale of computational expense is determined by LLM inference, a small ensemble of probe models can be trained without significantly increasing the overall scale of computational requirements.

Models within an ensemble are trained with different batch orderings of a random sample that is drawn without replacement from the dataset test split. The ensemble can also incorporate different label descriptions, different prompts, or other variations. Regardless of the variations within the ensemble, the most confident model from the ensemble is selected to make predictions. The constraint term of the probe loss can be maximized over the ensemble to search for the most confident model, but in our evaluations, we saw slightly improved performance by selecting the model maximizing $C_{confidence}$ for prediction:
\begin{equation}
    C_{confidence} = \sum{|P(x_+)-0.5|} 
\end{equation}

\subsection{Label translation}

Many datasets contain arbitrary labels that don't convey much semantic information. Since probes rely on semantic contrast in the latent vector space associated with the labels, arbitrary labels such as ``a" vs ``b" or even ``yes" vs ``no" don't convey as much useful semantic content as ``positive" vs ``negative" or ``good weather" vs ``bad weather". For the best performance, we ideally want labels that are descriptive and self-contained. Since datasets are under no obligation to provide such labels, we perform a translation step in which either a human or instruction-tuned language model answers the question ``What does (label 0) mean here?" to produce meaningful labels. The descriptive labels are substituted for the original labels. In the few-shot case, label occurences in the prompt should also be substituted with the new labels. This translation step needs to be done just once per task, during dataset creation, so human labling is fast and straightforward. Automatic description is also feasible, we evaluate generating this using the prompts `\textless base prompt \textgreater Describe the meaning of the response ``\textless class label \textgreater" to this question in three words or less.', for each class label. More robust and sophisticated appoaches are possible, the automatic label translation demonstrated in this study should be regarded as a proof-of-concept.

Semantically rich label descriptions can boost model performance significantly for some tasks. This tends to be the case when translated labels are descriptive, as concise as possible without losing detail, and semantically relevant to the query. Binary classification tasks admitting such a description are the types of tasks that could perform well with Glia methods - notably, the same label description should apply to every example in the dataset. Labels should be semantically equivalent across the dataset, not merely syntactically equivalent. The methods are better suited to true binary classification tasks, sorting examples into two meaningful equivalence classes, compared to arbitrary two-choice questions.

\subsection{Symmetry Breaking}

We propose a pretraining and training technique to break the probe's symmetry and enable fully unsupervised inference by associating labels with inputs. Without symmetry breaking, label orientation is randomly assigned during training. The model loss function has permuation symmetry - swapping $x_-$ for $x_+$ results in an identical loss. As such, there are two identical minima with equally good but opposite solutions, the one we reach is determined by the random initialization of the neural network parameters at the beginning of the training process. We break this symmetry before probe training, when we perform a few epochs of cross-entropy training of the probe model, without the final softmax layer, using the hidden states of candidate labels, an empty prompt, and/or examples from a few-shot prompt.

We implement two variants of symmetry breaking training, one designed for zero-shot inference use cases, and the other for few-shot use cases (or zero-shot cases where both labels appear in the prompt). In zero-shot mode, the data used for pretraining are hidden layer activations of the labels, obtained via a forward pass of the language model on the labels themselves - $\phi(-)$ and $\phi(+)$. We wish to assign these states the numerical indices 0 and 1, respectively. These are used to construct a two-element synthetic dataset in which the data are $(\phi(-),0)$ and $(\phi(+),1)$. The model is then pretrained using cross-entropy loss on these data for a few epochs. When multiple class labels appear in the prompt (e.g. in a few-shot prompt), this weakens the influence of directly associating labels with their hidden states. In order to ensure correlations across the ensemble, cross-entropy pretraining proceeds using the task prompt instead of the class labels, but substituted with an empty string (``") in place of an example. Positive and negative labels are appended to the prompt as before, the resulting pretraining data is $(\phi(``"_-),0)$ and $(\phi(``"_+),1)$. Additionally, when few-shot examples are present, we perform standard supervised cross-entropy training on the hidden states of the few-shot examples and their labels from the few-shot prompt.

Cross-entropy pretraining for several epochs 'spontaneously' breaks the symmetry\footnote{It is possible to make an analogy to spontaneous symmetry breaking in physical systems, such as permanent magnets. Without symmetry breaking, our model behaves like a ferromagnet that has lost its magnetization by being heated above the Curie temperature to a paramagnetic state in which electron spins are randomly oriented \citep{curie1895properties}. In this analogy, electron spins in a magnet are analogous to our label orientations. After symmetry is broken, the ensemble behaves like a ferromagnet below the Curie temperature, were electron spins have aligned with an external field, analogous to label orientations aligning across the ensemble.} of the ensemble, which undergoes a phase transition to a symmetry-broken state, with correlations arising among trained probes in the ensemble. This transition can be observed in figure 2. Finally, we note that pretraining is a statistical solution that can occasionally fail, resulting in models that do not have the preferred orientation. 
\begin{figure}[t]
   \centering
   \includegraphics[width=\textwidth]{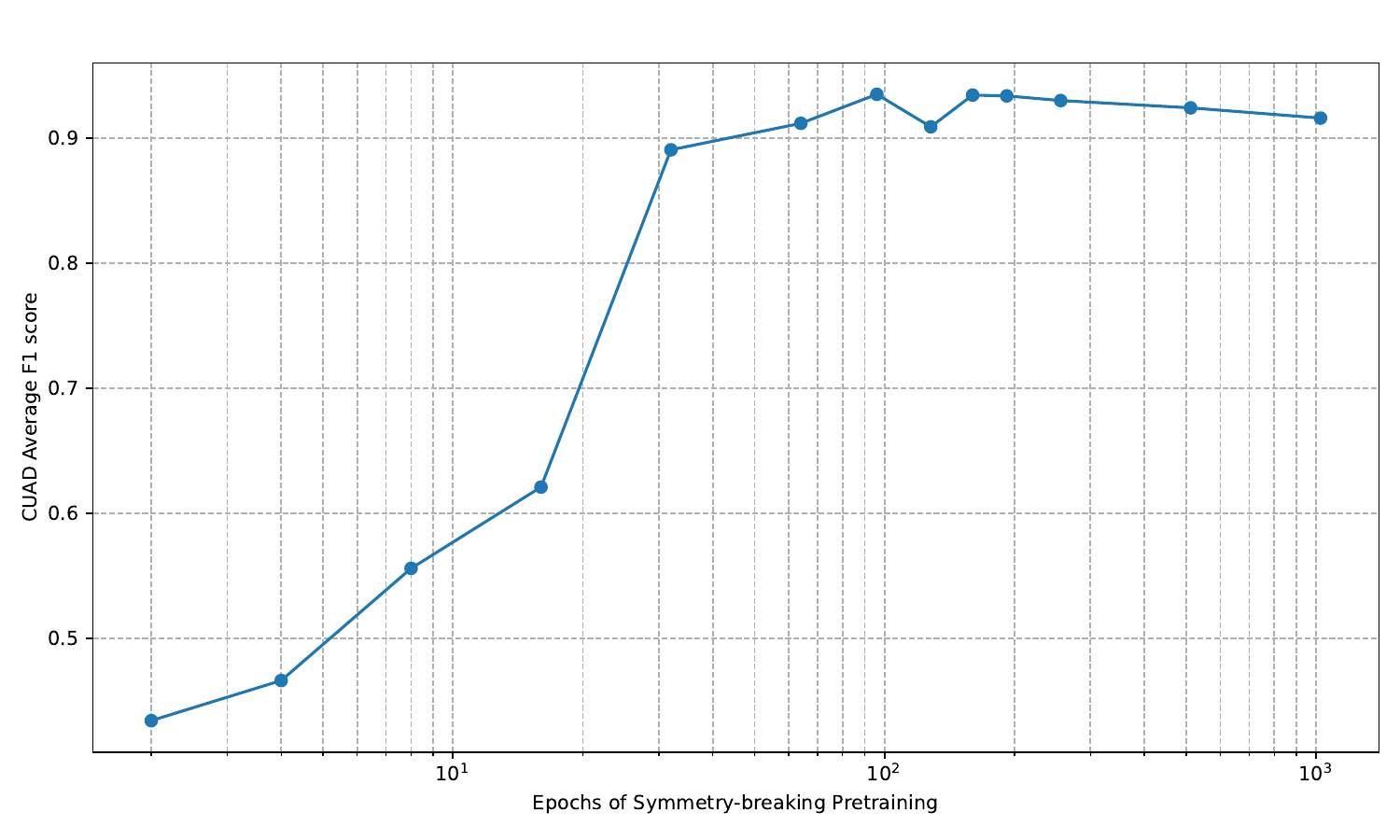}
   \caption{An illustration of spontaneous symmetry breaking in pretrained models. Glia inference is evaluated after n steps of symmetry-breaking pretraining, where n is shown in logarithmic scale on the x-axis. The y-axis shows average F1 score on the CUAD dataset for each n. A phase transition from a symmetric, disordered state to an asymmetric, ordered state is observed. Note that once symmetry has been broken, further pretraining slowly degrades performance.
   }
   \label{fig:ssb}
\end{figure}

\section{Evaluation}

We evaluate Glia on three types of tasks: sentiment classification using IMDB and Amazon reviews, legal contract understanding with CUAD (Contract Understanding Atticus Dataset), and legal issue identification with the croudsourced Learned Hands dataset. These contain a total of 49,089 examples spanning 56 tasks.

We perform two Glia evaluations. One evaluates the orientation of models in a semi-supervised manner, post-hoc, using evaluation data, akin to \citet{burns2023}. The other evaluation uses symmetry breaking. The symmetry breaking evaluation is fully unsupervised where prompts use zero-shot inference (IMDB and Amazon datasets). Symmetry breaking evaluation is is technically semi-supervised in the case of few-shot prompts since labeled examples are contained in the prompt, but no additional labels from the test dataset are used to make predictions.

We train ensembles of probe models for 4000 epochs using the Adam optimizer \citep{kingma2014adam} with a learning rate of $1 \times 10^{-6}$ and weight decay of 0.01. For symmetry breaking evaluation, we perform 128 epochs of symmetry-breaking pretraining. 9 models are trained per ensemble\footnote{We chose an odd ensemble size to facilitate voting, but found that winner-takes-all performed better than voting.}, and the most confident model is selected ($\arg \max C_{confidence}$). Unless otherwise specified, no training/test split is used. Loss hyperparameters are $\alpha=0.7$ and $\beta=0.3$.  

Evaluations are performed using five different LLMs as base models. We evaluate four decoder-only models - Llama-3-8B, Llama-3-8B-Instruct \citep{llama3modelcard}, Mistral-7B-v0.3, and Mistral-7B-v0.3-Instruct \citep{jiang2023mistral7b}. We also evaluate an encoder-only model, Deberta-v3-large \citep{he2021deberta, he2021debertav3}, to demonstrate the applicability of these methods to encoder models. For benchmarks, we evaluate three of the most capable LLMs currently available: OpenAI's GPT-4o, Google's Gemini-1.5-Pro, and Meta's Llama-3.1-405B-Instruct. We also benchmark the predecessor technique, CCS, using the same base LLMs as Glia. We benchmark calibrated zero-shot inference \citep{zhao2021} and calibrated few-shot inference using the decoder-only base models Mistral-7B-v0.3 and Llama-3-8B. We also train supervised probes on the 6-shot examples (training set) of the CUAD and Learned Hands tasks, without training using the unsupervised probe loss. Sentiment classification tasks use 0-shot rather than few-shot prompts and do not have access to labeled data, so they aren't benchmarked with supervised probes. Finally, we estimate an upper bound on probe performance ("Supervised ceiling") by splitting the test dataset in half and training a supervised probe model on half of the split. The performance of this supervised ``ceiling" probe on the remaining half of the test set is reported. This supervised ceiling is not a proper benchmark, it is evaluated on half of the test dataset in order to provide a rough estimate of the maximum possible probe performance. \citet{burns2023} similarly uses logistic regression to estimate a supervised upper bound for CCS. 

We evaluate with human-generated labels for the base models Mistral-7B-v0.3 and Llama-3-8B, and evaluate with machine-generated labels on the corresponding instruction-tuned chat models, Mistral-7b-Instruct-v0.3 and Llama-3-8B-Instruct. For the encoder-only model Deberta-v3-large, we do not perform automatic generation of labels (encoder models don't generate text) or perform zero-shot/few-shot baselines (we aren't fine-tuning the encoder-only model to make predictions). We use the 6-shot base prompts for CUAD and Learned Hands tasks and 0-shot inference for Amazon and IMDB sentiment analysis tasks. The Mistral-7B-v0.3 and Llama-3-8B benchmarks report calibrated zero-shot performance \citep{zhao2021} or calibrated few-shot performance, as appropriate. CCS methods use published parameter settings \citep{burns2023}.

On average, Glia+Mistral models performs better than Glia+Llama models, but in all cases Glia offered significant performance improvement over the base LLMs. Performance using Deberta was weaker than with the decoder-only models, as expected, although Glia+Deberta exceeded the larger Mistral 7B and Llama 8B models in terms of average zero-shot/few-shot performance. 

\subsection{Summary of results}

Table 1 shows the results of Glia evaluation using five different base LLMs. Mistral7B-v0.3-Instruct and Llama3-8B-Instruct are instructed to automaticaly generate the descriptive labels used for evaluation. For Mistral7B-v0.3, Llama-3-8B, and Deberta-v3-large, two human-generated descriptive labels (see Appendix) are provided for each task.
\begin{table*}[t]
	\centering
    \small
    \begin{tabular}{l|ccc|c}
	Method                             & Sentiment analysis & CUAD               & Learned Hands & Average (F1) \\
   \hline 
   GPT-4o                             & 95.8               & 86.5               & 84.5          & 88.9 \\
   Gemini-1.5-Pro                     & 95.4               & 91.4               & 81.4          & 89.4 \\
   Llama-3.1-405B-Instruct    & \textbf{96.4}              & 92.0               & 87.6          & 92.0 \\
   Mistral-7B-v0.3                    & 80.6               & 53.9               & 56.1          & 63.5 \\
   Llama-3-8B                         & 74.4               & 52.2               & 49.1          & 58.6 \\
   CCS+Mistral7B-v0.3                 & 85.4               & 69.0               & 54.6          & 69.7 \\
   CCS+Llama3-8B                      & 57.6               & 51.2               & 52.1          & 53.6 \\
   CCS+Deberta-v3-Large               & 69.7               & 51.5               & 54.8          & 58.7 \\
   Supervised probe - M7B             & -                & 80.2           & 67.2      & - \\
   Supervised probe - L8B             & -                & 59.3           & 52.3      & - \\
   Supervised probe - M7B-I           & -                & 56.9           & 57.6      & - \\
   Supervised probe - L8B-I           & -                & 58.4           & 51.3      & - \\
   Supervised probe - DEB             & -                & 56.9           & 43.3      & - \\ 
   \hline
   Supervised ceiling - M7B           & 96.5               & 95.3               & 96.5          & 96.1 \\
   Supervised ceiling - L8B           & 97.3               & 91.8               & 97.2          & 95.4 \\
   Supervised ceiling - M7B-I         & 96.2               & 95.8               & 98.5          & 96.8 \\
   Supervised ceiling - L8B-I         & 96.5               & 95.0               & 99.0          & 96.8 \\
   Supervised ceiling - DEB           & 95.3               & 86.2               & 95.3          & 92.2 \\ 
   \hline 
   Glia(post-hoc, M7B)                 & 95.7               & 93.1               & \textbf{88.1} & \textbf{92.3} \\
   Glia(post-hoc, L8B)                 & 95.5               & 79.8               & 78.3          & 84.5 \\
   Glia(post-hoc, auto, M7B-I)         & 95.4               & 91.3               & 80.5          & 89.1 \\
   Glia(post-hoc, auto, L8B-I)         & 95.4               & 88.9               & 69.7          & 84.7 \\
   Glia(post-hoc, DEB)                 & 95.3               & 72.0               & 49.9          & 72.4 \\ 
   Glia(broken symmetry, M7B)         & 95.5       & \textbf{93.4}              & 87.8          & 92.2 \\ 
   Glia(broken symmetry, L8B)         & 95.5               & 75.9               & 51.2          & 74.2 \\ 
   Glia(broken symmetry, auto, M7B-I) & 95.2               & 81.7               & 78.7          & 85.2 \\ 
   Glia(broken symmetry, auto, L8B-I) & 95.4               & 63.4               & 77.9          & 78.9 \\ 
   Glia(broken symmetry, DEB)         & 95.2               & 60.4               & 46.3          & 67.3 \\ 
   
    \end{tabular}
	\caption{
	F1 scores for all models averaged across all tasks for each dataset. Glia+M7B, Glia+L8B, and Glia+DEB refer to Glia methods paired with Mistral-7B-v0.3, Llama-3-8B, and Deberta-v3-large, respectively. Glia(auto, M7B-I) and Glia(auto) models automatically generate descriptive labels. Glia models paired with non-instruction-tuned base LLMs use human-described labels. Glia(post-hoc) methods are semi-supervised and use dataset labels to determine label orientation. The Glia(broken symmetry) methods use symmetry-breaking pretraining to orient labels, they are fully unsupervised for zero-shot prompts and semi-supervised for few-shot prompts, only having access to labeled data that appears in the few-shot prompt. Supervised probes are trained only on the few-shot examples, supervised ceiling is an upper bound trained on half of the test data. Glia models use an ensemble size of 9. Symmetry-breaking here is model-dependent, Glia models using Mistral 7B models take 32 epochs of symmetry-breaking pretraining, Glia models using Llama 8B take 64 epochs, and Glia using Deberta takes 1024 epochs.
    }
    \label{tab:summary}
    \vspace{-1pt}
\end{table*}
We randomly select 10,000 reviews from IMDB (\verb|stanfordnlp/imdb| dataset) and 10,000 reviews from Amazon \verb|amazon_polarity| dataset) for sentiment classification. These are the easiest tasks in the evaluation, with the performance of several models clustered around 95-96 percent. This level of performance is significantly better than CCS and zero-shot inference using the same base LLMs, however. We also evaluate the CUAD dataset \citep{hendrycks2021cuad} using prompts from Legalbench \citep{guha2023legalbench}. This consists of 38 tasks and 17980 total examples, tasks involve determining whether legal contracts contain particuar types of clauses.  The Learned Hands dataset croudsources labels for legal issues identified in online posts. Labels are determined through a game-like voting process in which users flag legal issues in documents. Learned Hands is the most difficult dataset in our evaluation. We evaluate performance on the Learned Hands dataset using prompts from Legalbench \citep{guha2023legalbench}. This comprises 16 tasks and 11109 total examples. Gemini-1.5-Pro's safety filter blocked 216 of the examples in this dataset, these examples are omitted from the evaluation. Llama-3-405B blocked 13 of 10,000 IMDB reviews. Given their relatively low proportion, these blocked examples do not affect the results. On the CUAD and Learned Hands datasets, Glia using Mistral-7B-v0.3 as a base model scores higher on average than the baseline models GPT-4o, Gemini-1.5-Pro, and Llama-3-405B. On sentiment classification tasks, Glia performed similarly to the larger baseline models. 

\subsection{Ablation testing}

We also perform ablation testing for the varous techniques introduced by this study, with the exception of symmetry breaking, which was effectively ablated (vs. post-hoc label orientations) in the main evaluation. For ablation testing, we evaluate ablations post-hoc using the CUAD dataset and two base LLMs, Mistral 7B 0.3 and Llama-3-8B, to show model dependence. Results are shown in Table \ref{tab:ablate}. In the loss ablation test, the technique is evaluated using the CCS loss function from \cite{burns2023} but with all the other Glia techniques active and non-ablated. In the translation ablation test, labels simply aren't translated, and the ensemble ablation test uses an ensemble size of 1. Ablating individual component techniques does not account for all of the performance gains observed from Glia inference. We believe this is due to interactions between the Glia methods. The importance of various components depends on LLM and task performance. When LLM performance is good on a particular task, some of these techniques may not be impactful, but when LLM performance is bad, they may all be necessary to get an optimal result.

\begin{table*}[t]
	\centering
    \small
    \begin{tabular}{l|cc}
	Ablation & Glia+Mistral7B-v0.3 & Glia+Llama3-8B \\
   \hline
   Ensemble      & 93.1        & 70.4 \\
   Loss function & 92.1        & 69.6 \\       
   Label translation   & 86.5        & 75.7 \\ 
   \hline
   Glia CUAD baseline & \textbf{93.4} & \textbf{80.1} \\
   Calibrated 5-shot (no Glia) & 53.9 & 52.2 \\  
   \hline
    \end{tabular}
	\caption{
	Ablation test results for Glia methods on the CUAD dataset, using Mistral-7B-v0.3 and Llama-3-8B v0.3 as base models.
    }
    \label{tab:ablate}
    \vspace{-1pt}
\end{table*}

\subsection{Computational resources}
GPT-4o and Gemini-1.5-Pro were evaluated through their respective APIs, and Llama 3.1 405B was hosted by Fireworks.AI. All other testing, including all Glia LLM inference and probe training, was performed using a single NVIDIA RTX 4090 GPU. Detailed information regarding the computational resources of the baseline models are not available, but are believed to exceed this by multiple orders of magnitude. Simply based on parameter count, Llama 3's 405 billion parameters are more than 55 times the size of Glia+Mistral v0.3's 7.3 billion parameters, with proportional hardware reqirements. 

In contrast to 0-shot inference requirig one forward pass through an LLM, Glia requires two forward passes, in addition to training the ensemble of probes. In the evaluations presented here, training ensembles of 9 models takes roughly the same amount of time as the two forward passes, so the entire procedure takes roughly 4x the amount of compute as 0-shot inference. We note that single-model ensembles achieved competitive performance using the Mistral 7B v0.3 base model and human-generated labels, in that case the computational requirement is roughly twice that of 0-shot inference. Based on the 55x performance gain and a 4x increase in computational requirements, we believe the Glia methods represent an order-of-magnitude increase in efficiency.
\section{Discussion and Conclusion}
We have seen that the Glia methods can dramatically improve LLM inference across a wide range of binary classification tasks. Inference performance can be comparable to LLMs requiring orders of magnitude more computational resources. In our evaluation, the Mistral 7B models performed better than the Llama 8B models, both in terms of correctly orienting labels and probe performance, in spite of these LLMs being roughly comparable in size and capability. The reasons for this discrepancy are not currently well understood, and additional research into the suitability of particular LLMs for these methods would be beneficial.

A possibly significant direction for future development is the extension of these methods to other types of constraints. More general constraints could extend the applicability of the method from binary classification to multiclass classification and potentially benefit generative use cases. Constraints based on prior knowledge could potentially mitigate the hallucination problem to make generative AI more accurate and trustworthy. 

All evaluations in this study used text-based examples. Evaluating multimodal binary classification performance of the Glia techniques could potentially demonstrate applicability to cover a broader range of task modalities. To explore the applicability of Glia methods to general document analysis, evaluating longer context (100k+ tokens) tasks could be beneficial. The examples used in this study have fewer than 8192 tokens, evaluating and/or extending the method with longer contexts could enable practical use cases for longer documents.

\subsubsection*{Author Contributions}
J. Scoville contributed, implemented, and evaluated the methods introduced in this paper. S. Gao selected evaluation datasets and performed evaluations. D. Agrawal suggested the incorporation of supervised learning and provided general feedback on the manuscript. J. Qadrud-Din performed evaluations and provided technical advice and guidance throughout the study.

\bibliography{glia-iclr}
\bibliographystyle{modified_style}

\appendix
\section{Appendix}
\subsection{Task Prompts}
Where logprobs are available, we directly compare logprobs for the two classes to perform 0-shot and few-shot inference. For externally hosted models where logprobs are not readily available (GPT-4o, Gemini-1.5-Pro, Meta-Llama-405B), system prompts instruct the model to answer only with the relevant class labels. For example: \\
\verb|"Answer questions with only the word 'positive' or the word | \\
\verb|'negative', and nothing further at all. No additional text or | \\
\verb|punctuation."| \\

\subsubsection{IMDB}
\verb|Consider this movie review: {{text}} Between "negative" | \\
\verb|and "positive", the sentiment of this review is |
\subsubsection{Amazon}
\verb|"Consider this Amazon product review: {{text}} The sentiment of | \\
\verb|this review is "|
\subsubsection{CUAD and Learned Hands}
CUAD and Learned Hands tasks use their 6-shot base prompts from Legalbench. \citep{guha2023legalbench} \\ 
(\verb|https://github.com/HazyResearch/legalbench/|)

\subsection{Labels}
For automatic label translation, we first extract the question from each base prompt for each task by truncating any few-shot examples from the base prompt. This text is inserted into the label translation prompt, along with one of the original task labels:\\
\verb|<Question> Describe the meaning of the response "<label>" to this|
\verb|question in three words or less.|

The human-provided labels for each task can be seen in table \ref{tab:labels} and \ref{tab:labels2}. Note that many of these simply rephrase the task name.

\begin{table*}[t]
	\centering
    \small
    \begin{tabular}{l|c|c}
	Task name & Label & Label description \\
   \hline
   amazon\_polarity & positive & positive \\
   amazon\_polarity & negative & negative \\ 
   stanfordnlp/imdb & pos & positive \\
   stanfordnlp/imdb & neg & negative \\
   cuad\_affiliate\_license-licensee    & Yes & Affiliate license \\
   cuad\_affiliate\_license-licensee & No & No affiliate license \\
   cuad\_affiliate\_license-licensor    & Yes & License by affiliate \\
   cuad\_affiliate\_license-licensor    & No & Affiliate unlicensed \\
   cuad\_anti-assignment & Yes & Consent or notice required for assignment \\
   cuad\_anti-assignment & No & Consent or notice not required for assignment \\
   cuad\_audit\_rights & Yes & Right to audit \\
   cuad\_audit\_rights & No & No right to audit \\
   cuad\_cap\_on\_liability & Yes & Cap on liability \\
   cuad\_cap\_on\_liability & No & No cap on liability \\
   cuad\_change\_of\_control & Yes & Change of control rights \\
   cuad\_change\_of\_control & No & No change of control rights \\
   cuad\_competitive\_restriction\_exception & Yes & Competitive restriction exceptions \\
   cuad\_competitive\_restriction\_exception & No & No competitive restriction exceptions \\
   cuad\_covenant\_not\_to\_sue & Yes & Covenant not to sue \\
   cuad\_covenant\_not\_to\_sue & No & No Covenant not to sue \\
   cuad\_effective\_date & Yes & Effective date is specified \\
   cuad\_effective\_date & No & Effective date is not specified \\
   cuad\_exclusivity & Yes & Exclusivity \\
   cuad\_exclusivity & No & No exclusivity \\
   cuad\_expiration\_date & Yes & Expriation date specified \\
   cuad\_expiration\_date & No & Not specified \\
   cuad\_governing\_law & Yes & Yes \\
   cuad\_governing\_law & No & No \\
   cuad\_insurance & Yes & Insurance requirement \\
   cuad\_insurance & No & No insurance requirement \\
   cuad\_ip\_ownership\_assignment & Yes & Intellectual property rights are assigned \\
   cuad\_ip\_ownership\_assignment & No & Intellectual property rights are not assigned \\
cuad\_irrevocable\_or\_perpetual\_license & Yes & Irrevocable or perpetual license grant specified \\
cuad\_irrevocable\_or\_perpetual\_license & No & No irrevocable or perpetual license grant specified \\
cuad\_joint\_ip\_ownership  & Yes & Joint IP ownership specified \\
cuad\_joint\_ip\_ownership & No & No Joint IP ownership specified \\
cuad\_license\_grant & Yes & License granted \\
cuad\_license\_grant & No & No license granted \\
cuad\_liquidated\_damages  & Yes & Liquidated damages awarded \\
cuad\_liquidated\_damages & No & No liquidated damages awarded \\
cuad\_minimum\_commitment & Yes & Minimum specified \\
cuad\_minimum\_commitment & No & No minimum specified \\
cuad\_most\_favored\_nation & Yes & Entitled to better terms, most favored nation \\
cuad\_most\_favored\_nation & No & Not entitled to better terms or most favored nation \\
cuad\_no-solicit\_of\_customers & Yes & No-solicit of customers restriction \\
cuad\_no-solicit\_of\_customers & No & No no-solicit of customers restriction \\
cuad\_no-solicit\_of\_employees & Yes & No-solicit of employees restriction \\
cuad\_no-solicit\_of\_employees & No & No no-solicit of employees restriction \\
cuad\_non-compete & Yes & Non-competition is specified \\
cuad\_non-compete & No & Non-competition is not specified \\
cuad\_non-disparagement & Yes & Non-disparagement is specified \\
cuad\_non-disparagement & No & Non-disparagement is not specified \\
cuad\_non-transferable\_license & Yes & Non-transferable license specified \\
cuad\_non-transferable\_license & No & Non-transferable license not specified \\
cuad\_notice\_period\_to\_terminate\_renewal  & Yes & Notice period to terminate renewal specified \\
cuad\_notice\_period\_to\_terminate\_renewal & No & Notice period to terminate renewal not specified \\
cuad\_post-termination\_services & Yes & Post-termination services \\
cuad\_post-termination\_services & No & No post-termination services \\

\hline
\end{tabular}
\caption{
Human-generated descriptive labels used for evaluation of Glia with Mistral-7B-v0.3, Llama-3-8B, and Deberta-v3-large. Part 1 of 2.
}
\label{tab:labels}
\vspace{-1pt}
\end{table*}

\begin{table*}[t]
	\centering
    \small
    \begin{tabular}{l|c|c}
	Task name & Label & Translated label \\
   \hline
   cuad\_price\_restrictions & Yes & Restriction on the ability to raise or reduce prices \\
cuad\_price\_restrictions & No & No restriction on the ability to raise or reduce prices \\
cuad\_renewal\_term & Yes & Renewal term specified \\
cuad\_renewal\_term & No & Renewal term not specified \\
cuad\_revenue-profit\_sharing & Yes & Revenue-profit sharing required \\
cuad\_revenue-profit\_sharing & No & No revenue-profit sharing required \\
cuad\_rofr-rofo-rofn & Yes & Right of first refusal, right of first offer, \\ 
 & & or right of first negotiation granted \\
cuad\_rofr-rofo-rofn & No & No right of first refusal, right of first offer, \\
 & & or right of first negotiation granted \\
cuad\_source\_code\_escrow & Yes & Source code escrow requried \\
cuad\_source\_code\_escrow & No & Source code escrow not required \\
cuad\_termination\_for\_convenience & Yes & Termination for convenience \\
cuad\_termination\_for\_convenience & No & No termination for convenience \\
cuad\_third\_party\_beneficiary & Yes & Third party beneficiary specified \\
cuad\_third\_party\_beneficiary & No & No third party beneficiary specified \\
cuad\_uncapped\_liability & Yes & Uncapped liability specified \\
cuad\_uncapped\_liability & No & Uncapped liability not specified \\
cuad\_unlimited-all-you-can-eat-license & Yes & Unlimited-all-you-can-eat license granted \\
cuad\_unlimited-all-you-can-eat-license & No & No unlimited-all-you-can-eat license granted \\
cuad\_volume\_restriction & Yes & Volume restriction specified \\
cuad\_volume\_restriction & No & No volume restriction specified \\
cuad\_warranty\_duration & Yes & Warranty duration specified \\
cuad\_warranty\_duration & No & Warranty duration not specified \\
learned\_hands\_benefits & Yes & Disusses benefits \\
learned\_hands\_benefits & No & Doesn't discuss benefits \\
learned\_hands\_business & Yes & Discusses business \\
learned\_hands\_business & No & Doesn't discuss business \\
learned\_hands\_consumer & Yes & Consumer issues \\
learned\_hands\_consumer & No & No consumer issues \\
learned\_hands\_courts & Yes & Court issues \\
learned\_hands\_courts & No & No court issues \\
learned\_hands\_crime & Yes & Criminal issues \\
learned\_hands\_crime & No & No criminal issues \\
learned\_hands\_divorce & Yes & Divorce issues \\
learned\_hands\_divorce & No & No divorce issues \\
learned\_hands\_domestic\_violence & Yes & Domestic violence \\
learned\_hands\_domestic\_violence & No & No domestic violence \\
learned\_hands\_education & Yes & Educational issues \\
learned\_hands\_education & No & No educational issues \\
learned\_hands\_employment & Yes & Employment issues \\
learned\_hands\_employment & No & No employment issues \\
learned\_hands\_estates & Yes & Estate issues \\
learned\_hands\_estates & No & No estate issues \\
learned\_hands\_family & Yes & Family issues \\
learned\_hands\_family & No & No family issues \\
learned\_hands\_health & Yes & Health related \\
learned\_hands\_health & No & Not health related \\
learned\_hands\_housing & Yes & Housing related \\
learned\_hands\_housing & No & Not housing related \\
learned\_hands\_immigration & Yes & Immigration related \\
learned\_hands\_immigration & No & Not immigration related \\
learned\_hands\_torts & Yes & Civil liablity \\
learned\_hands\_torts & No & No civil liability \\
learned\_hands\_traffic & Yes & Traffic issues \\
learned\_hands\_traffic & No & No traffic issues \\
   \hline
    \end{tabular}
	\caption{
	Human-generated descriptive labels used for evaluation of Glia with Mistral-7B-v0.3, Llama-3-8B, and Deberta-v3-large. Part 2 of 2.
   }
    \label{tab:labels2}
    \vspace{-1pt}
\end{table*}

\end{document}

%% file: math_commands.tex
\usepackage{amsmath,amsfonts,bm}

\def\eqref#1{equation~\ref{#1}}

\def\1{\bm{1}}

\def\rx{{\textnormal{x}}}

\DeclareMathAlphabet{\mathsfit}{\encodingdefault}{\sfdefault}{m}{sl}
\SetMathAlphabet{\mathsfit}{bold}{\encodingdefault}{\sfdefault}{bx}{n}

